\documentclass{article}
\usepackage{ijcai26}

\usepackage{times}
\usepackage{soul}
\usepackage{url}
\usepackage[hidelinks]{hyperref}
\usepackage[utf8]{inputenc}
\usepackage[small]{caption}
\usepackage{graphicx}
\usepackage{amsmath}
\usepackage{amsthm}
\usepackage{booktabs}
\usepackage{algorithm}
\usepackage{algorithmic}
\usepackage[switch]{lineno}
\usepackage{amssymb}
\usepackage{multirow}
\usepackage{subcaption}
\usepackage{placeins}
\usepackage{pgffor}
\usepackage{graphicx}

\usepackage{enumitem} %
\usepackage[table]{xcolor}

\title{FedCARE: Federated Unlearning with Conflict-Aware Projection and Relearning-Resistant Recovery}

\begin{document}
\author{
    Yue Li$^1$,
    Mingmin Chu$^1$,
    Xilei Yang$^1$,
    Da Xiao$^1$,
    Ziqi Xu$^2$,
    Wei Shao$^{3,4,2,5}$,
    Qipeng Song$^1$,
    Hui Li$^1$
    \affiliations
    $^1$School of Cyber Engineering, Xidian University\\
    $^2$RMIT University\\
    $^3$Commonwealth Scientific and Industrial Research Organisation (CSIRO)\\
    $^4$UNSW Sydney\\
    $^5$University of California, Davis
    \emails
    \{liyue, xiaoda, qpsong\}@xidian.edu.cn, 
    \{24151213797, 24151213670\}@stu.xidian.edu.cn, \\
    ziqi.xu@rmit.edu.au,
    phdweishao@gmail.com,
    lihui@mail.xidian.edu.cn
}
\maketitle

\begin{abstract}
Federated learning (FL) enables collaborative model training without centralizing raw data, but privacy regulations such as the right to be forgotten require FL systems to remove the influence of previously used training data upon request. Retraining a federated model from scratch is prohibitively expensive, motivating federated unlearning (FU). However, existing FU methods suffer from high unlearning overhead, utility degradation caused by entangled knowledge, and unintended relearning during post-unlearning recovery. In this paper, we propose \textbf{FedCARE}, a unified and low overhead FU framework that enables conflict-aware unlearning and relearning-resistant recovery. FedCARE leverages gradient ascent for efficient forgetting when target data are locally available and employs data free model inversion to construct class level proxies of shared knowledge. Based on these insights, FedCARE integrates a pseudo-sample generator, conflict-aware projected gradient ascent for utility preserving unlearning, and a recovery strategy that suppresses rollback toward the pre-unlearning model. FedCARE supports client, instance, and class level unlearning with modest overhead. Extensive experiments on multiple datasets and model architectures under both IID and non-IID settings show that FedCARE achieves effective forgetting, improved utility retention, and reduced relearning risk compared to state of the art FU baselines. 

\end{abstract}

\section{Introduction}
\paragraph{Background.}

Federated Learning (FL) enables collaborative model training without centralizing raw data by aggregating locally computed updates from distributed clients~\cite{mcmahan2017communication}. In practice, an FL system must not only continuously improve the model as new data arrive, but also be able to eliminate the influence of previously used training data upon request in order to comply with privacy regulations. For instance, the European Union’s General Data Protection Regulation (GDPR)~\cite{voigt2017eu} establishes the right to erasure, often referred to as the ``right to be forgotten'' (RTBF). A straightforward way to comply with RTBF is to remove the target data and retrain the federated model from scratch; however, this approach is often prohibitively expensive in terms of computational resources. Inspired by machine unlearning in centralized settings~\cite{bourtoule2021machine}, Federated Unlearning (FU)~\cite{liu2021federaser} has emerged as a promising alternative, aiming to remove a target client’s (or a data subset’s) contribution from a federated global model without retraining from scratch.

\paragraph{Challenges.}

Despite encouraging progress (refer to Sec.~\ref{sec:related_work} for more details), practical FU still faces the following challenges: \textbf{(1) Low-cost and unified unlearning.} A deployable FU framework should complete as quickly as possible upon an unlearning request, while incurring only a small incremental resource cost and minimally affecting non-target clients. Moreover, real unlearning requests arise at different granularities (client-, sample-, and class-level), which calls for a unified design that can cover diverse unlearning paradigms. \textbf{(2) Utility degradation due to knowledge entanglement.}Unlearning updates are often computed from a limited view of the data (e.g., only the requesting client), while the global model encodes entangled client-specific and global shared knowledge. As a result, removing a specific contribution can inadvertently damage shared knowledge, leading to noticeable utility drops on retained data, especially under non-IID client distributions.\textbf{(3) Relearning during post-unlearning recovery.} To recover model utility, FL commonly continues training on remaining clients after unlearning.  Yet this recovery phase can drift the model back toward its pre-unlearning state, implicitly restoring removed knowledge and weakening the unlearning effect.

\paragraph{Method.}

To address the above challenges, we propose \textbf{FedCARE}, a novel FU framework that achieves \textbf{C}onflict-\textbf{A}ware unlearning and \textbf{RE}learning-resistant recovery to retain model utility. The design of FedCARE is motivated by two key observations: (1) when the data to be forgotten are available at the target client, gradient ascent provides a simple and low-cost unlearning primitive, provided that its collateral damage to the remaining knowledge can be effectively constrained; and (2) data-free synthesis via model inversion can recover informative class-level signals from a trained model, which can serve as a proxy for shared knowledge and help constrain utility degradation during unlearning without requiring access to real reference data.

Concretely, FedCARE follows a three-stage procedure. First, the server trains a lightweight class-conditional pseudo-sample generator once in a data-free manner and makes it available for future unlearning requests. Second, taking client-level unlearning as an example, the target client uses the generator to synthesize pseudo-samples, computes a reference gradient and performs conflict-aware projected gradient ascent to achieve utility-preserving unlearning. Third, during recovery, the remaining clients freeze their local backbone (feature extractor) and update only the classifier head, while the server filters aggregated updates along a relearning direction to suppress rollback toward the pre-unlearning model. Finally, FedCARE also supports instance- and class-level unlearning by adjusting the target dataset and the corresponding reference gradient, enabling unified unlearning with low overhead.

\paragraph{Contributions.}




Our contributions are as follows:
\begin{itemize}[leftmargin=*,noitemsep,topsep=2pt]
    \item We propose \textbf{FedCARE}, a low-overhead FU framework that supports client-, instance-, and class-level unlearning while reducing resource overhead and minimizing impact on non-target clients.
    
    \item We introduce a conflict-aware projected forgetting mechanism that constrains forgetting updates with a utility-related protection reference to mitigate utility degradation.
    
    \item We design a relearning-resistant recovery strategy that suppresses rollback during post-unlearning recovery via client-specific backbone freezing and server-side update filtering.
    
    \item Extensive experiments on multiple datasets and architectures under both IID and non-IID settings show that FedCARE achieves effective forgetting, improved utility retention, and reduced relearning risk with modest additional overhead compared to state-of-the-art FU baselines.
\end{itemize}



\section{Related Work}
\label{sec:related_work}

Federated Unlearning (FU) integrates federated learning with machine unlearning, aiming to enable privacy-preserving data removal in distributed environments. Traditional centralized unlearning methods are unsuitable for such settings due to their reliance on global access to training data. In FU, client data remains local and is never shared with other clients or the central server; instead, only locally computed model updates are transmitted for aggregation into a global model.

Directly deleting target data is infeasible due to the memorization of sensitive information in trained models~\cite{nasr2018comprehensive,melis2019exploiting,song2020analyzing}. Existing federated unlearning approaches can be broadly categorized into storage-based and storage-free methods, both of which face critical limitations in practice. Storage-based methods~\cite{liu2021federaser,DBLP:conf/ijcai/LinGDNGCR24} accelerate retraining by retaining historical model updates, but incur substantial storage overhead that scales with training duration and client participation, limiting their deployment in long-running or resource-constrained systems. Storage-free methods avoid historical state retention but often sacrifice efficiency or robustness: MoDe~\cite{zhao2024MoDe} relies on knowledge distillation from randomly degraded models, introducing optimization stochasticity and high communication cost; NoT~\cite{DBLP:conf/cvpr/KhalilBL0B025} employs heuristic weight negation that severely disrupts learned representations and requires extensive fine-tuning to recover utility; and FedOSD~\cite{DBLP:conf/aaai/PanWLZW0Z25} mitigates interference via orthogonal projection but depends on continuous client participation for complex gradient computations, exacerbating computational overhead. Collectively, these limitations reveal the lack of a unified federated unlearning framework that can achieve effective forgetting with low overhead, preserve model utility, and remain robust to unintended relearning during recovery.

\begin{figure*}[t]
    \centering
    \includegraphics[width=0.9\textwidth]{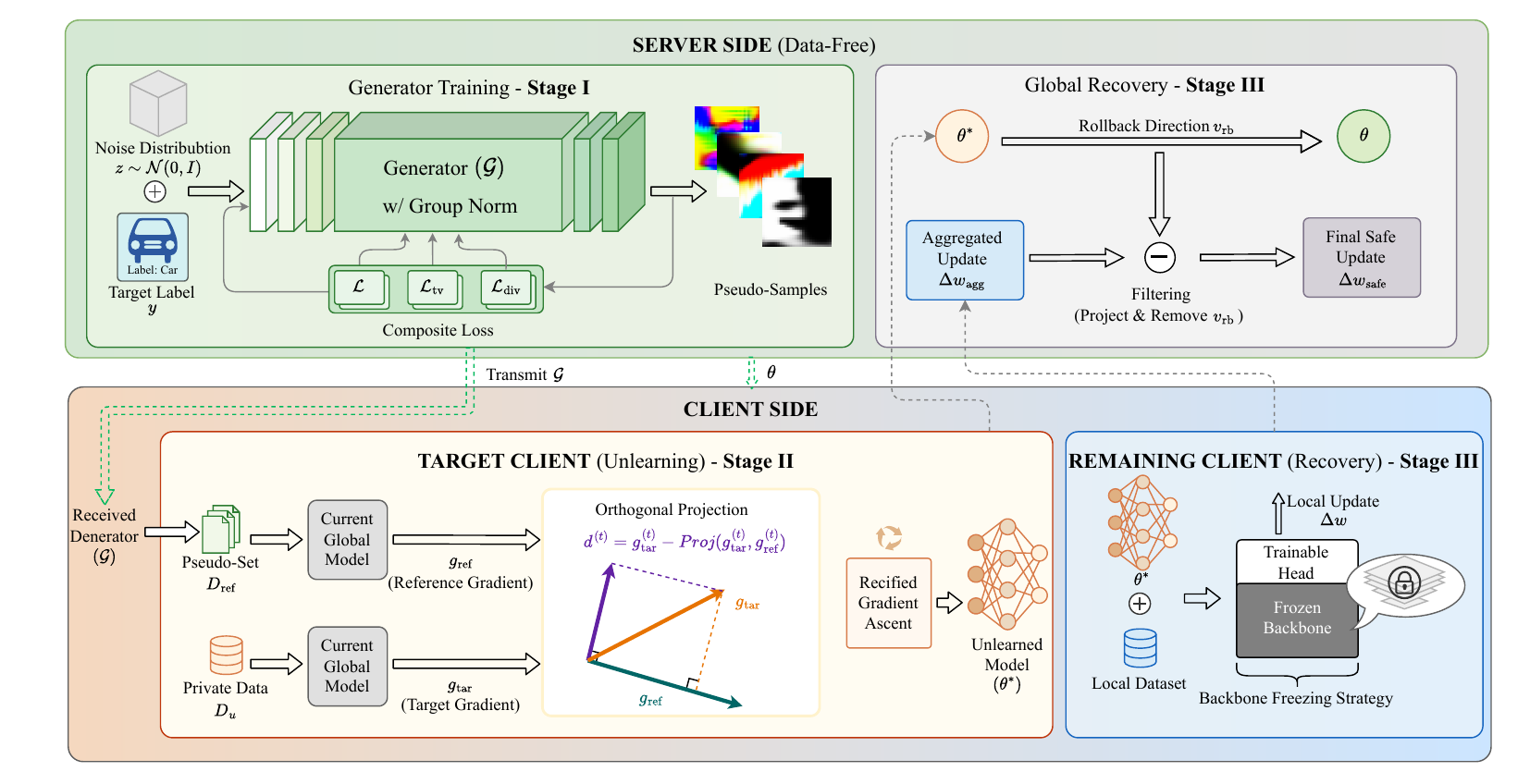}
    \caption{Overall framework illustrated with client-level unlearning. After federated training, the server trains a lightweight data-free generator once. Upon an unlearning request, the generator is sent to the target client, which performs conflict-aware projection unlearning using its private forget set and pseudo-samples. The remaining clients then conduct constrained recovery with relearning-resistant aggregation.}
    \label{fig:framework}
\end{figure*}

\section{Methodology}

In this section, we introduce \textbf{FedCARE}, which consists of three main stages: training a data-free pseudo-sample generator, performing conflict-aware projection unlearning, and executing relearning-resistant model recovery. The overall architecture is illustrated in Figure~\ref{fig:framework}.

\subsection{Data-Free Pseudo-Sample Generator} 



FedCARE requires a lightweight, data-free utility reference to support constrained unlearning. Since the server cannot access any real retained data, we construct a proxy reference set $\mathcal{D}_{\mathrm{ref}}$ by synthesizing class-conditional pseudo-samples from the trained global model $\theta$. Importantly, this generator is trained once at the server and can be reused for future unlearning requests. Synthesized examples and generator training cost are provided in the supplementary material.

We implement $\mathcal{G}$ as a lightweight decoder-style network that maps a latent code and a class label to the input space. Given $z \sim \mathcal{N}(0, I)$ and a label $y$, we first obtain an initial low-resolution feature map $h_0 = \mathrm{Proj}(z, y)$ through a learnable projection layer. The features are then progressively refined and upsampled via a cascade of $K$ decoder blocks:
\begin{equation}
    \tilde{x} = \mathcal{G}(z, y) = \phi\!\left((B_K \circ B_{K-1} \circ \cdots \circ B_1)\big(h_0\big)\right),
    \label{eq:generator_arch}
\end{equation}
where $\phi(\cdot)$ denotes an output activation function (e.g., $\tanh$ or sigmoid) to ensure that the generated samples lie within the valid input range. Each decoder block $B_i(\cdot)$ increases the spatial resolution while refining semantic structure, thereby producing class-consistent pseudo-samples.

Prior data-free synthesis methods, such as DeepInversion~\cite{DBLP:conf/cvpr/YinMALMHJK20}, typically rely on Batch Normalization (BN) to regularize the generation process by matching stored feature statistics. However, in non-IID federated settings, the BN statistics aggregated at the server often suffer from severe distribution shifts and fail to accurately reflect the true global data distribution~\cite{hsieh2020non}. Relying on such corrupted statistics can lead to the generation of semantic artifacts rather than valid, generic features. To address this issue, we replace BN with Group Normalization (GN)~\cite{DBLP:conf/eccv/WuH18} in our generator blocks:
\begin{equation}
    B_i(h) = \sigma \Big( \mathrm{GN} \big( \mathrm{Conv}_i ( \mathrm{Up}_i(h) ) \big) \Big),
    \label{eq:block_def}
\end{equation}
where $\mathrm{Up}(\cdot)$ denotes nearest-neighbor upsampling and $\sigma(\cdot)$ is the activation function. Since GN normalizes features independently of batch statistics, it decouples the generation process from unreliable global distributions, ensuring that the synthesized samples are guided solely by the semantic constraints imposed by the model inversion objective.

To ensure that the constructed feature subspace effectively approximates the generic features of globally shared knowledge, we formulate a unified optimization objective for data-free synthesis as follow:
\begin{equation}
    \mathcal{L}_{\mathrm{total}} = \mathcal{L} + \lambda_{\mathrm{tv}}\mathcal{L}_{\mathrm{tv}} + \lambda_{\mathrm{div}}\mathcal{L}_{\mathrm{div}},
    \label{eq:specific_losses}
\end{equation}
where each term plays a distinct role in shaping the synthesized pseudo-samples. We next describe the design and motivation of these components in detail.

To anchor the pseudo-sample $\tilde{x}$ within the correct decision boundary, we adopt the standard cross-entropy loss with respect to the target class $y$. To mitigate high-frequency artifacts inherent to data-free synthesis~\cite{odena2016deconvolution}, we further incorporate an anisotropic total variation regularizer,
\begin{equation}
\mathcal{L}_{\mathrm{tv}} = \sum_{i,j} \big( |x_{i+1,j} - x_{i,j}| + |x_{i,j+1} - x_{i,j}| \big),
\end{equation}which imposes a smoothness prior on the generated samples. 

Moreover, to prevent mode collapse and encourage diverse generations, we introduce a diversity regularization term $\mathcal{L}_{\mathrm{div}}$, defined as the inverse of the feature distance ratio:
\begin{equation}
\mathcal{L}_{\mathrm{div}} = \Big[ \frac{\| \mathcal{G}(z_1, y) - \mathcal{G}(z_2, y) \|_1}{\| z_1 - z_2 \|_1} + \epsilon \Big]^{-1}.
\end{equation}

This term enforces a sensitivity constraint on the generator, ensuring that distinct latent codes produce sufficiently different feature realizations and thereby promoting adequate coverage of the latent subspace.

To mitigate gradient domination caused by scale disparity between loss terms (i.e., $\mathcal{L}_{\mathrm{tv}} \gg \mathcal{L}_{\mathrm{ce}}$), we adopt a hybrid loss balancing strategy with adaptive initialization of the smoothness weight $\lambda_{\mathrm{tv}}$. At the start of training ($t = 0$), $\lambda_{\mathrm{tv}}$ is set as $\lambda_{\mathrm{tv}} = \eta \cdot (\mathcal{L}_{\mathrm{ce}}^{(0)} / \mathcal{L}_{\mathrm{tv}}^{(0)})$, where $\eta \in [10^{-3}, 10^{-1}]$ is an empirical attenuation factor inversely related to the dataset’s textural complexity. This normalization aligns the gradient magnitudes of the smoothness and semantic objectives, preventing early stage optimization bias. In contrast, we assign a fixed constant weight to the diversity term, setting $\lambda_{\mathrm{div}}$ to a small value (e.g., $0.5$) to maintain a stable repulsive force for feature space exploration.

\subsection{Conflict-Aware Projection Unlearning}
\label{sec：ca-unlearning}


For clarity, we take client-level unlearning as an illustrative example. At the onset of unlearning, the server shares the trained pseudo-sample generator $\mathcal{G}$ with the target client $u$. Let $\theta$ denote the current global model parameters and $\mathcal{D}_{u}$ the private dataset to be forgotten.

A simple and low-cost primitive for client unlearning is gradient ascent on $\mathcal{D}_{u}$, which maximizes the empirical loss to erase the client-specific contribution. However, directly maximizing $\mathcal{L}_{u}(\theta) \triangleq \mathcal{L}(\theta;\mathcal{D}_{u})$ can severely degrade model utility, as client-specific and globally shared representations are entangled in $\theta$. To address this issue, we formulate unlearning as a constrained optimization problem:
\begin{equation}
\max_{\Delta \theta}\ \mathcal{L}_{u}(\theta+\Delta \theta)
\quad \mathrm{s.t.}\quad
\mathcal{L}_{\mathrm{ref}}(\theta+\Delta \theta)\ \le\ \mathcal{L}_{\mathrm{ref}}(\theta),
\label{eq:constrained_ul}
\end{equation}
where $\mathcal{L}_{\mathrm{ref}}(\theta)$ denotes a reference loss that serves as a proxy for retained utility. In practice, $\mathcal{L}_{\mathrm{ref}}$ cannot be evaluated on real retained data due to data isolation and privacy constraints. FedCARE overcomes this limitation by constructing a proxy reference dataset $\mathcal{D}_{\mathrm{ref}}$ using the generator $\mathcal{G}$ (for class-level unlearning, $\mathcal{D}_{\mathrm{ref}}$ excludes the forgotten class), and defining the reference loss as $\mathcal{L}_{\mathrm{ref}}(\theta) \triangleq \mathcal{L}(\theta;\mathcal{D}_{\mathrm{ref}})$.


Directly solving Eq.~\eqref{eq:constrained_ul} is intractable for deep neural networks. We therefore enforce the utility constraint locally using a first-order approximation. Let $d$ denote the ascent direction and consider an update of the form $\Delta\theta = \eta d$ with a small step size $\eta$. Applying a first-order Taylor expansion to the reference loss yields:
\begin{equation}
\mathcal{L}_{\mathrm{ref}}(\theta + \eta d)
\approx
\mathcal{L}_{\mathrm{ref}}(\theta) + \eta \left\langle \nabla_{\theta}\mathcal{L}_{\mathrm{ref}}(\theta),\, d \right\rangle.
\label{eq:taylor}
\end{equation}


Thus, a sufficient condition to prevent an increase in the reference loss at first order is:
\begin{equation}
\left\langle g_{\mathrm{ref}},\, d \right\rangle \le 0,
\quad \text{where} \quad
g_{\mathrm{ref}} \triangleq \nabla_{\theta}\mathcal{L}_{\mathrm{ref}}(\theta).
\label{eq:local_constraint}
\end{equation}

Meanwhile, the unconstrained ascent direction for forgetting is given by:
\begin{equation}
g_{\mathrm{tar}} \triangleq \nabla_{\theta}\mathcal{L}_{u}(\theta).
\label{eq:gtar}
\end{equation}



%

Given the unconstrained ascent direction $g_{\mathrm{tar}}$ and the first-order utility constraint in Eq.~\eqref{eq:local_constraint}, the problem reduces to finding a feasible update direction that remains as close as possible to $g_{\mathrm{tar}}$ while preserving the reference loss.

In practice, computing $g_{\mathrm{ref}}$ using the entire proxy dataset $\mathcal{D}_{\mathrm{ref}}$ at every iteration is computationally unnecessary. Moreover, the local constraint in Eq.~\eqref{eq:local_constraint} is defined with respect to the current parameters $\theta^{(t)}$, and the associated gradient field evolves as the model is updated. To account for this dynamics while maintaining efficiency, we estimate the reference gradient online using a mini-batch $\mathcal{B}_{\mathrm{ref}}^{(t)} \subset \mathcal{D}_{\mathrm{ref}}$ at each step $t$:
\begin{equation}
g_{\mathrm{ref}}^{(t)} \triangleq \nabla_{\theta}\mathcal{L}\!\left(\theta^{(t)};\mathcal{B}_{\mathrm{ref}}^{(t)}\right),
\label{eq:gref_batch}
\end{equation}
which serves as a lightweight and step-adaptive proxy for retained utility. Similarly, the forgetting gradient is computed on a mini-batch $\mathcal{B}_{u}^{(t)} \subset \mathcal{D}_{u}$:
\begin{equation}
g_{\mathrm{tar}}^{(t)} \triangleq \nabla_{\theta}\mathcal{L}\!\left(\theta^{(t)};\mathcal{B}_{u}^{(t)}\right).
\label{eq:gtar_batch}
\end{equation}


At each step $t$, we obtain the closest feasible ascent direction to $g_{\mathrm{tar}}^{(t)}$ by solving the following constrained optimization problem:
\begin{equation}
d^{(t)}
= \arg\min_{d}\ \left\|d - g_{\mathrm{tar}}^{(t)}\right\|_2^2
\quad \mathrm{s.t.}\quad
\left\langle g_{\mathrm{ref}}^{(t)},\, d \right\rangle \le 0,
\label{eq:qp_step}
\end{equation}
which projects $g_{\mathrm{tar}}^{(t)}$ onto the half-space defined by the local utility constraint. This problem admits a closed-form solution given by:
\begin{equation}
d^{(t)}
= g_{\mathrm{tar}}^{(t)} -
\frac{\max\!\left(0, \left\langle g_{\mathrm{tar}}^{(t)},\, g_{\mathrm{ref}}^{(t)} \right\rangle \right)}
{\left\| g_{\mathrm{ref}}^{(t)} \right\|_2^2 + \epsilon}\, g_{\mathrm{ref}}^{(t)},
\label{eq:hinge_proj}
\end{equation}
where $\epsilon$ is a small constant introduced for numerical stability. Intuitively, when $\langle g_{\mathrm{tar}}^{(t)}, g_{\mathrm{ref}}^{(t)} \rangle \le 0$, the unconstrained ascent direction already satisfies the utility constraint and remains unchanged. Otherwise, the minimal conflicting component along $g_{\mathrm{ref}}^{(t)}$ is removed so that the corrected direction lies exactly on the constraint boundary.


Finally, the target client updates the model parameters via projected gradient ascent for $T$ steps:
\begin{equation}
\theta^{(t+1)} \leftarrow \theta^{(t)} + \eta_{\mathrm{ul}}\, d^{(t)},
\qquad t = 0, \ldots, T-1,
\label{eq:update}
\end{equation}
where $\eta_{\mathrm{ul}}$ denotes the unlearning learning rate.

\subsection{Relearning-resistant Model Recovery}
\label{sec:recovery}

Although the projection mechanism in Sec.~3.2 constrains utility degradation during the erasure process, the performance on retained tasks may still degrade in practice. Consequently, FU methods commonly adopt a post unlearning recovery stage, such as standard FedAvg fine tuning, by continuing federated training over the remaining clients’ local datasets, denoted by $\mathcal{D}_{\mathrm{rem}}$. However, such recovery procedures can inadvertently reintroduce partially removed knowledge, thereby weakening the effectiveness of unlearning~\cite{DBLP:conf/ccs/Chen000HZ21,DBLP:conf/aaai/PanWLZW0Z25}.



Following~\cite{DBLP:conf/iclr/IlharcoRWSHF23}, we adopt task vectors to represent weight space directions that encode task specific changes. Specifically, the knowledge removed during unlearning can be characterized as:
\begin{equation}
    \Delta\theta_{\mathrm{ul}} \triangleq \theta^{*} - \theta,
\end{equation}
where $\theta$ denotes the global model parameters immediately before unlearning and $\theta^{*}$ denotes the unlearned model. 

Similarly, the knowledge acquired during the recovery stage is given by:
\begin{equation}
    \Delta\theta_{\mathrm{rec}} \triangleq \theta^{*,r} - \theta^{*},
\end{equation}
where $\theta^{*,r}$ denotes the recovered model. Relearning, also referred to as rollback, becomes prominent when recovery updates move the model back toward $\theta$, that is, when $\Delta\theta_{\mathrm{rec}}$ contains a positive component along the rollback direction $\theta - \theta^{*} = -\Delta\theta_{\mathrm{ul}}$:
\begin{equation}
    \left\langle \Delta\theta_{\mathrm{rec}},\ -\Delta\theta_{\mathrm{ul}} \right\rangle > 0.
\label{eq:rollback-risk}
\end{equation}

Accordingly, a recovery strategy should explicitly suppress update components that align with the rollback direction.

FedCARE mitigates this effect through a dual client and server design. At the client side, we decompose the original and unlearned models as $\theta = (\phi, w)$ and $\theta^{*} = (\phi^{*}, w^{*})$, where $\phi$ and $w$ denote the backbone feature extractor and classifier head, respectively. During recovery, each remaining client freezes the backbone at $\phi^{*}$ and updates only the classifier head for a small number of rounds. This restricts recovery to adjusting decision boundaries within the post unlearning representation space, reducing the risk of reconstructing previously removed features.

At the server side, with the backbone fixed, recovery updates are confined to the classifier head subspace. Suppressing rollback in the full parameter space therefore reduces to preventing head updates from moving toward the original head $w$. Nevertheless, due to client data heterogeneity and optimization noise, the aggregated head update may still contain a non negligible rollback component. To address this, we further filter the aggregated head update along a fixed rollback direction in the head subspace.

Following the server side filtering strategy described above, we explicitly characterize the rollback direction in the classifier head subspace. We define the normalized rollback direction as:
\begin{equation}
    \mathbf{v}_{\mathrm{rb}} \triangleq \mathrm{Normalize}\!\left(w - w^{*}\right),
\end{equation}
where $w$ denotes the original classifier head and $w^{*}$ denotes the head after unlearning. Let $\Delta w_{\mathrm{agg}}$ be the standard aggregated head update from the remaining clients in a recovery round. Analogous to the conflict aware projection in Sec.~3.2, FedCARE computes the closest corrected update that does not move toward $w$ along $\mathbf{v}_{\mathrm{rb}}$, which admits the following closed form projection:
\begin{equation}
\Delta w_{\mathrm{safe}}
=
\Delta w_{\mathrm{agg}}
-
\frac{\max\!\left(0, \left\langle \Delta w_{\mathrm{agg}}, \mathbf{v}_{\mathrm{rb}} \right\rangle \right)}
{\left\|\mathbf{v}_{\mathrm{rb}}\right\|_2^2 + \epsilon}\cdot \mathbf{v}_{\mathrm{rb}}.
\end{equation}

This filtering guarantees that $\langle \Delta w_{\mathrm{safe}}, \mathbf{v}_{\mathrm{rb}} \rangle \le 0$, which eliminates first order rollback along the defined direction while still allowing effective utility recovery within the constrained parameter space.

\subsection{Instance- and Class-level Unlearning Support}
\label{subsec:extensions}
FedCARE supports different unlearning granularities with minimal modifications to the pipeline described in Secs.~\ref{sec：ca-unlearning}-\ref{sec:recovery}. When only a subset of samples on the target client needs to be removed, we define the forget set as $\mathcal{D}_f \subset \mathcal{D}_u$ and compute the forgetting update using $\mathcal{D}_f$, that is, by replacing $\mathcal{D}_u$ with $\mathcal{D}_f$ in the forget loss and gradient in Sec.~\ref{sec：ca-unlearning}. The utility reference construction via the pseudo-sample generator $\mathcal{G}$, the conflict-aware projection, and the subsequent recovery procedure remain unchanged.


For class-level unlearning of a class $c$, we apply two coordinated modifications. During unlearning, the forgetting update is computed using only local samples labeled $c$ on the target client (e.g., the client with the largest amount of class $c$ data). When constructing the utility reference, $\mathcal{G}$ synthesizes pseudo-samples only from the remaining label set excluding $c$, ensuring that the projection constraint does not preserve the removed class. Meanwhile, all remaining clients remove class $c$ samples from their local datasets and follow the same recovery protocol described in Sec.~\ref{sec:recovery}.




\section{Experiments}

\subsection{Experimental Setup}

\paragraph{Datasets, Models, and Experimental Settings.} We evaluate FedCARE on four standard datasets: MNIST~\cite{lecun1998gradient}, SVHN~\cite{netzer2011reading}, and CIFAR-10/100~\cite{alex2009learning}. For MNIST, we adopt a simple CNN consisting of two convolutional layers followed by two fully connected layers. For SVHN, CIFAR-10, and CIFAR-100, we use ResNet-18~\cite{he2016deep} as the backbone model. In federated learning, we employ the standard FedAvg algorithm for training with 10 clients under both IID and non-IID settings. For the non-IID case, data are partitioned according to a Dirichlet distribution $\mathrm{Dir}(\alpha)$ with concentration parameter $\alpha = 0.1$~\cite{DBLP:conf/icml/YurochkinAGGHK19}. Further implementation details are provided in the supplementary material.



\paragraph{Baselines.}
We compare FedCARE with six baselines, including Retraining and five state-of-the-art FU methods: FedEraser~\cite{liu2021federaser}, FedSE~\cite{DBLP:conf/ijcai/LinGDNGCR24}, MoDe~\cite{zhao2024MoDe}, FedOSD~\cite{DBLP:conf/aaai/PanWLZW0Z25}, and NoT~\cite{DBLP:conf/cvpr/KhalilBL0B025}. Note that FedEraser and FedSE are limited to client-level unlearning.


\paragraph{Evaluation Metrics.}
We evaluate the effectiveness and efficiency of FedCARE using the following metrics: {(1) Accuracy Metrics:} We use R-ACC to denote the classification accuracy on classes excluding the unlearning target, measuring the retention of non-target knowledge. U-ACC represents the accuracy on classes containing the forgotten data for assessing forgetting effectiveness, while Test denotes the overall accuracy of the global model on the complete test set. For U-Acc, R-Acc and Test, the ideal objective is to closely match the performance of the Retraining baseline. {(2) Membership Inference Attack (MIA)~\cite{DBLP:conf/csfw/YeomGFJ18}:} This metric quantifies privacy leakage by measuring the extent to which the forgotten dataset $\mathcal{D}_{u}$ can still be inferred as part of the training data. {(3) Attack Success Rate (ASR):} We employ ASR to assess whether implanted backdoor triggers have been fully removed from the model by injecting triggers into the target client’s data and flipping the corresponding labels following~\cite{DBLP:journals/corr/abs-2207-05521,DBLP:journals/corr/abs-1708-06733}. {(4) Communication and Computation Costs:} To evaluate system efficiency, we report the total time overhead incurred during the unlearning and recovery phases as the communication cost, along with the FLOPs~\cite{DBLP:journals/corr/abs-2001-08361} consumed by the remaining clients as the computation cost.

\subsection{Main Results}


\begin{table}[t!]
    \centering
    \scriptsize
    \setlength{\tabcolsep}{3pt}
    \renewcommand{\arraystretch}{0.7}
        \begin{tabular}{llcccccc}
            \toprule
            \multirow{2}{*}{} & \multirow{2}{*}{\textbf{Method}} & \multicolumn{3}{c}{\textbf{Accuracy}} & \multicolumn{1}{c}{\textbf{Privacy}} & \multicolumn{2}{c}{\textbf{Cost}} \\
            \cmidrule(lr){3-5} \cmidrule(lr){6-6} \cmidrule(lr){7-8}
             & & \textbf{R-Acc} & \textbf{U-Acc} & \textbf{Test}  & \textbf{MIA} ($\downarrow$) & \textbf{Time} ($\downarrow$) & \textbf{FLOPs} ($\downarrow$) \\
            \midrule
            
            \multirow{7}{*}{\rotatebox{90}{\shortstack{MNIST}}}
            & Retrain & 95.59\% & 94.43\% & 95.11\% & 50.21\% & 2441.20s & 7.67$\times10^{12}$ \\
            \cmidrule(lr){2-8}
            & FedEraser & 94.85\% & 96.11\% & 94.42\% & 56.16\% & 631.64s & 3.28$\times10^{12}$ \\
            & FedSE & 85.44\% & 72.10\% & 83.63\% & 63.80\% & 500.24s & 1.30$\times10^{12}$ \\
            & MoDe & 93.62\% & 90.55\% & 91.50\% & 53.72\% & 796.75s & 7.57$\times10^{11}$ \\
            & FedOSD & 95.05\% & 92.17\% & 93.81\% & 53.47\% & 180.20s & 4.12$\times10^{11}$ \\
            & NoT & 95.61\% & 94.70\% & 95.24\% & 54.24\% & 210.78s & 6.11$\times10^{11}$ \\
            & {FedCARE} & {95.11\%} & {93.80\%} & {94.91\%} & {50.18\%} & {54.41s} & 1.54$\times10^{11}$ \\
            \midrule

            \multirow{7}{*}{\rotatebox{90}{\shortstack{SVHN}}} 
            & Retrain & 90.98\% & 84.27\% & 86.37\% & 50.13\% & 7565.44s & 6.52$\times10^{15}$ \\
            \cmidrule(lr){2-8}
            & FedEraser & 90.14\% & 88.61\% & 86.16\% & 58.42\% & 1420.35s & 1.85$\times10^{15}$ \\
            & FedSE & 79.43\% & 68.25\% & 78.50\% & 63.83\% & 1150.28s & 2.68$\times10^{15}$ \\
            & MoDe & 85.20\% & 79.16\% & 84.89\% & 61.58\% & 2890.61s & 1.15$\times10^{15}$ \\
            & FedOSD & 91.88\% & 81.14\% & 85.21\% & 53.60\% & 385.49s & 6.90$\times10^{14}$ \\
            & NoT & 90.92\% & 83.33\% & 85.32\% & 54.10\% & 3410.10s & 2.38$\times10^{15}$ \\
            & {FedCARE} & {90.46\%} & {84.51\%} & {86.36\%} & {50.22\%} & {90.66s} & 2.07$\times10^{14}$ \\
            \midrule

            \multirow{7}{*}{\rotatebox{90}{\shortstack{CIFAR-10}}}
            & Retrain & 81.62\% & 78.63\% & 80.37\% & 50.28\% & 7892.47s & 5.66$\times10^{15}$ \\
            \cmidrule(lr){2-8}
            & FedEraser & 80.97\% & 75.15\% & 77.58\% & 57.30\% & 1466.31s & 2.25$\times10^{15}$ \\
            & FedSE & 66.45\% & 58.20\% & 65.10\% & 63.45\% & 1372.27s & 3.29$\times10^{15}$ \\
            & MoDe & 77.80\% & 62.50\% & 72.35\% & 60.54\% & 3365.68s & 1.49$\times10^{15}$ \\
            & FedOSD & 83.11\% & 73.26\% & 79.64\% & 54.10\% & 487.29s & 8.65$\times10^{14}$ \\
            & NoT & 81.58\% & 77.53\% & 79.33\% & 57.42\% & 4078.19s & 2.98$\times10^{15}$ \\
            & {FedCARE} & {81.30\%} & {78.52\%} & {80.01\%} & {50.38\%} & {125.39s} & 1.49$\times10^{14}$ \\
            \midrule

            \multirow{7}{*}{\rotatebox{90}{\shortstack{CIFAR-100}}}
            & Retrain & 61.98\% & 56.01\% & 60.03\% & 50.19\% & 15474.10s & 1.25$\times10^{16}$ \\
            \cmidrule(lr){2-8}
            & FedSE & 40.60\% & 32.17\% & 38.45\% & 66.10\% & 4159.66s & 3.13$\times10^{15}$ \\
            & MoDe & 47.88\% & 51.32\% & 46.20\% & 64.08\% & 5741.02s & 1.66$\times10^{15}$ \\
            & FedOSD & 62.40\% & 58.90\% & 61.85\% & 56.21\% & 873.95s & 1.05$\times10^{15}$ \\
            & NoT & 61.77\% & 55.46\% & 59.68\% & 50.15\% & 2955.30s & 2.67$\times10^{15}$ \\
            & {FedCARE} & {61.11\%} & {55.32\%} & {59.64\%} & {50.22\%} & {250.77s} & 4.07$\times10^{14}$ \\
            \bottomrule
        \end{tabular}
    
    \caption{{Client-level unlearning performance under non-IID settings.} We report results under non-IID scenarios to reflect more challenging and realistic unlearning conditions, and omit IID results where data redundancy tends to yield overly optimistic performance.}
    \label{tab:client_UL}
\end{table}

\begin{figure*}[t!]
    \centering
    \captionsetup[subfigure]{font=scriptsize,skip=3pt}
    \begin{subfigure}[b]{0.3\textwidth}
        \centering
        \includegraphics[width=\linewidth]{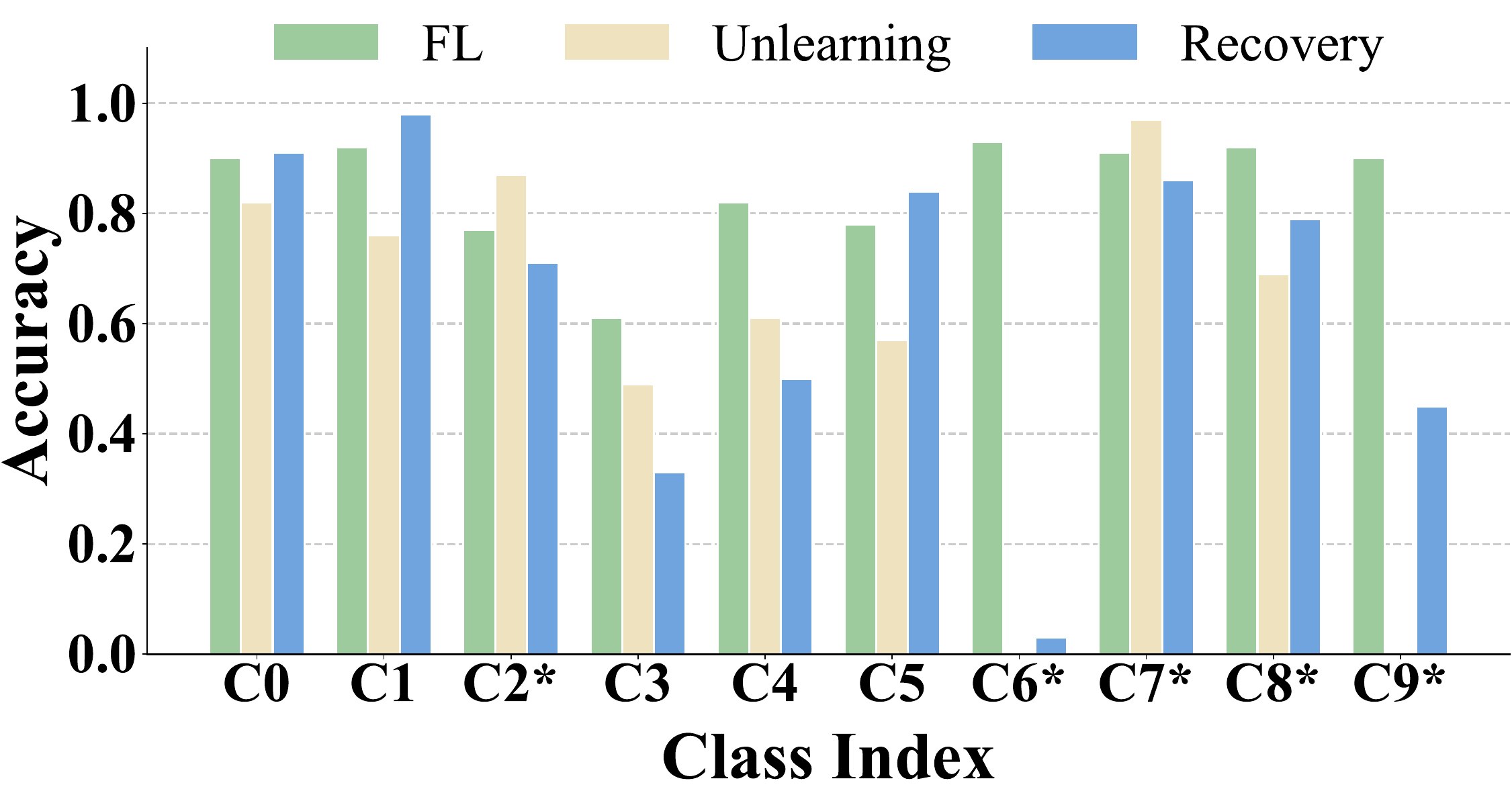}
        \caption{FedOSD}
        \label{fig:acc_client_FedOSD}
    \end{subfigure}
    \hfill 
    \begin{subfigure}[b]{0.3\textwidth}
        \centering
        \includegraphics[width=\linewidth]{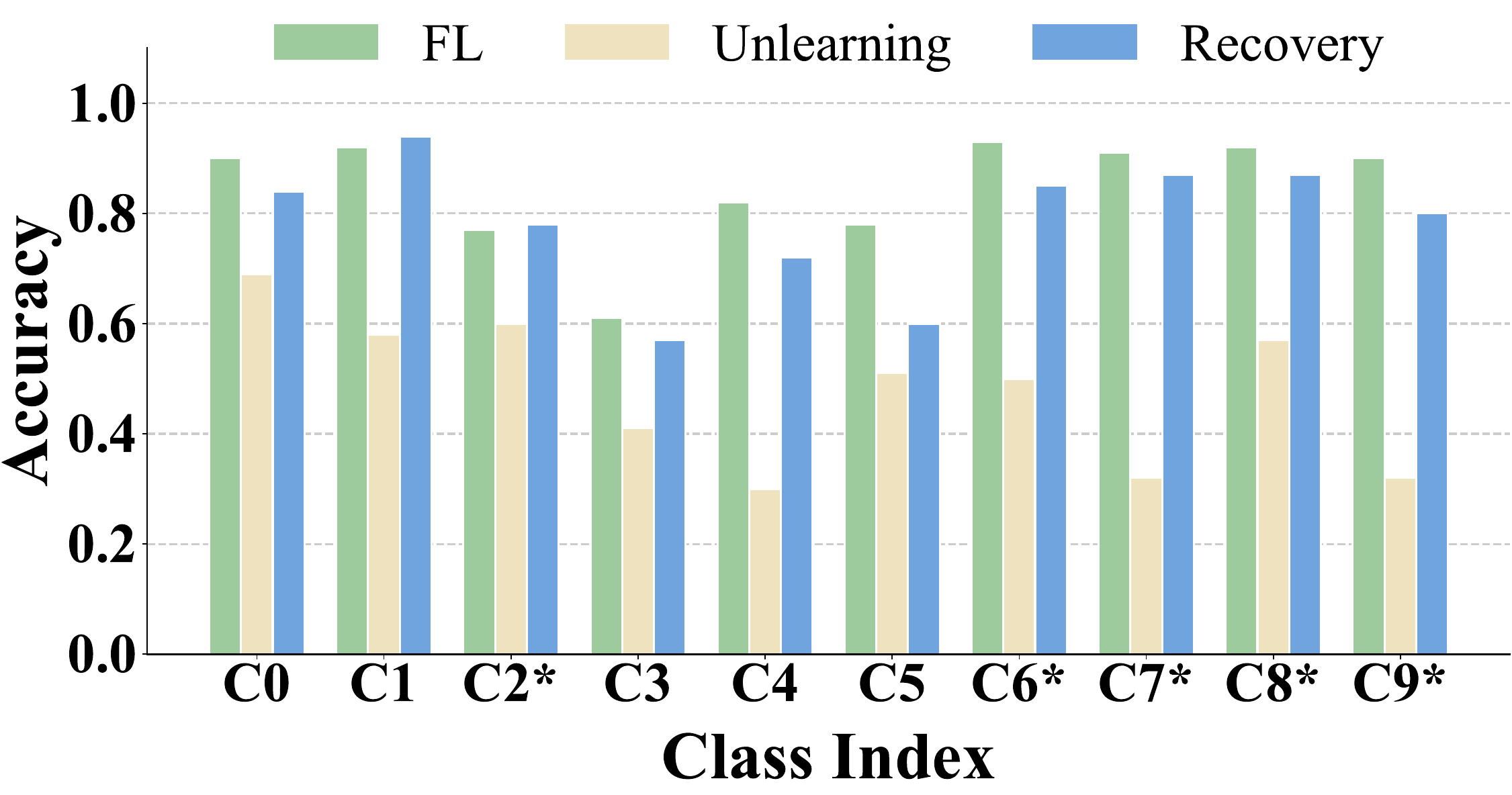}
        \caption{NoT}
        \label{fig:acc_client_NoT}
    \end{subfigure}
    \hfill 
    \begin{subfigure}[b]{0.3\textwidth}
        \centering
        \includegraphics[width=\linewidth]{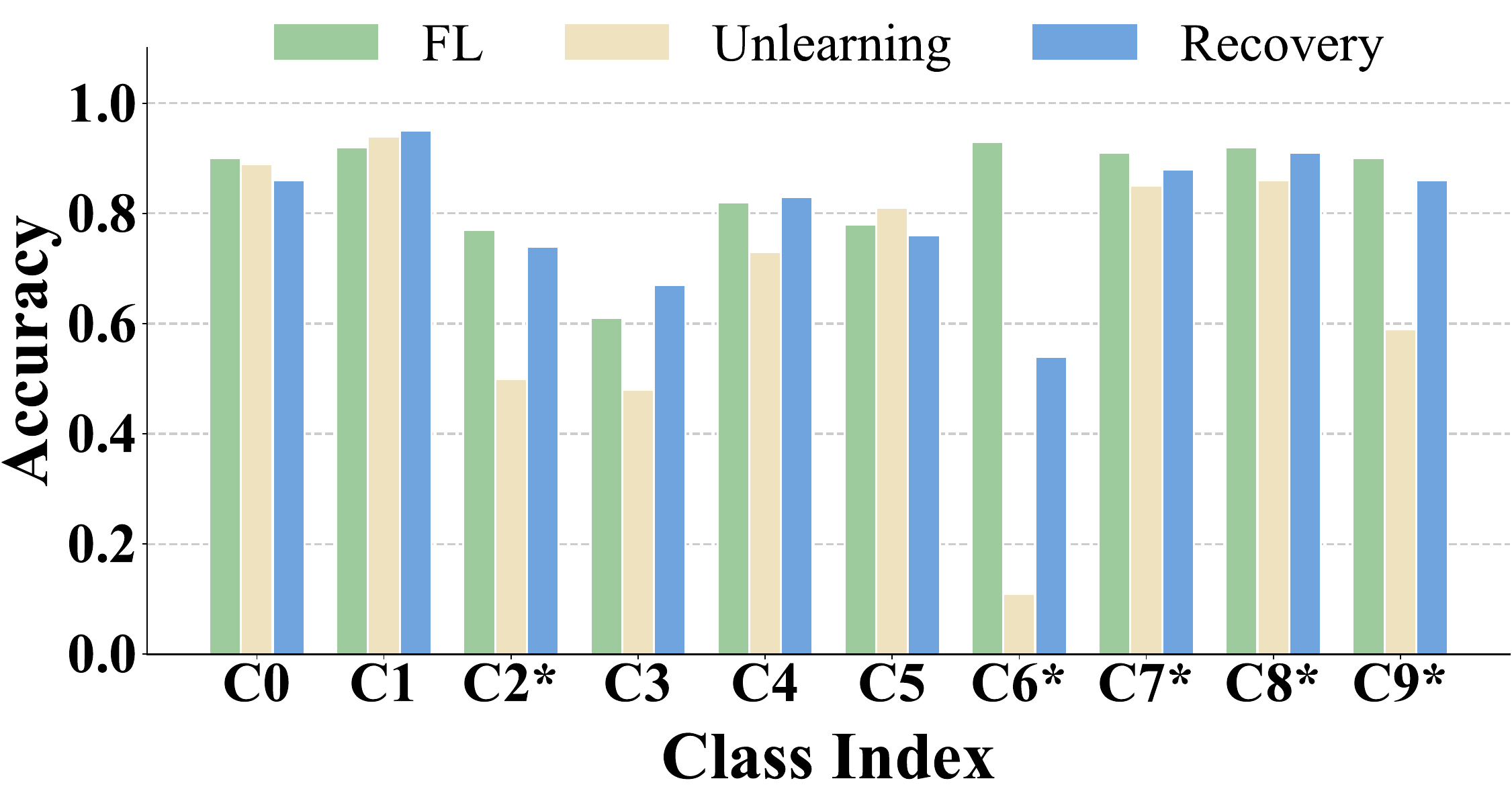}
        \caption{FedCARE}
        \label{fig:acc_client_FedCARE}
    \end{subfigure}
    
    \caption{{Client-level unlearning performance across different classes on CIFAR-10 (non-IID).} The ($*$) denotes classes contained within the unlearning client.}
    \label{fig:client_wise_accuracy}
\end{figure*}

\begin{figure*}[t!]
    \centering
    \captionsetup[subfigure]{font=scriptsize,skip=3pt}
    \begin{subfigure}[b]{0.3\textwidth}
        \centering
        \includegraphics[width=\linewidth]{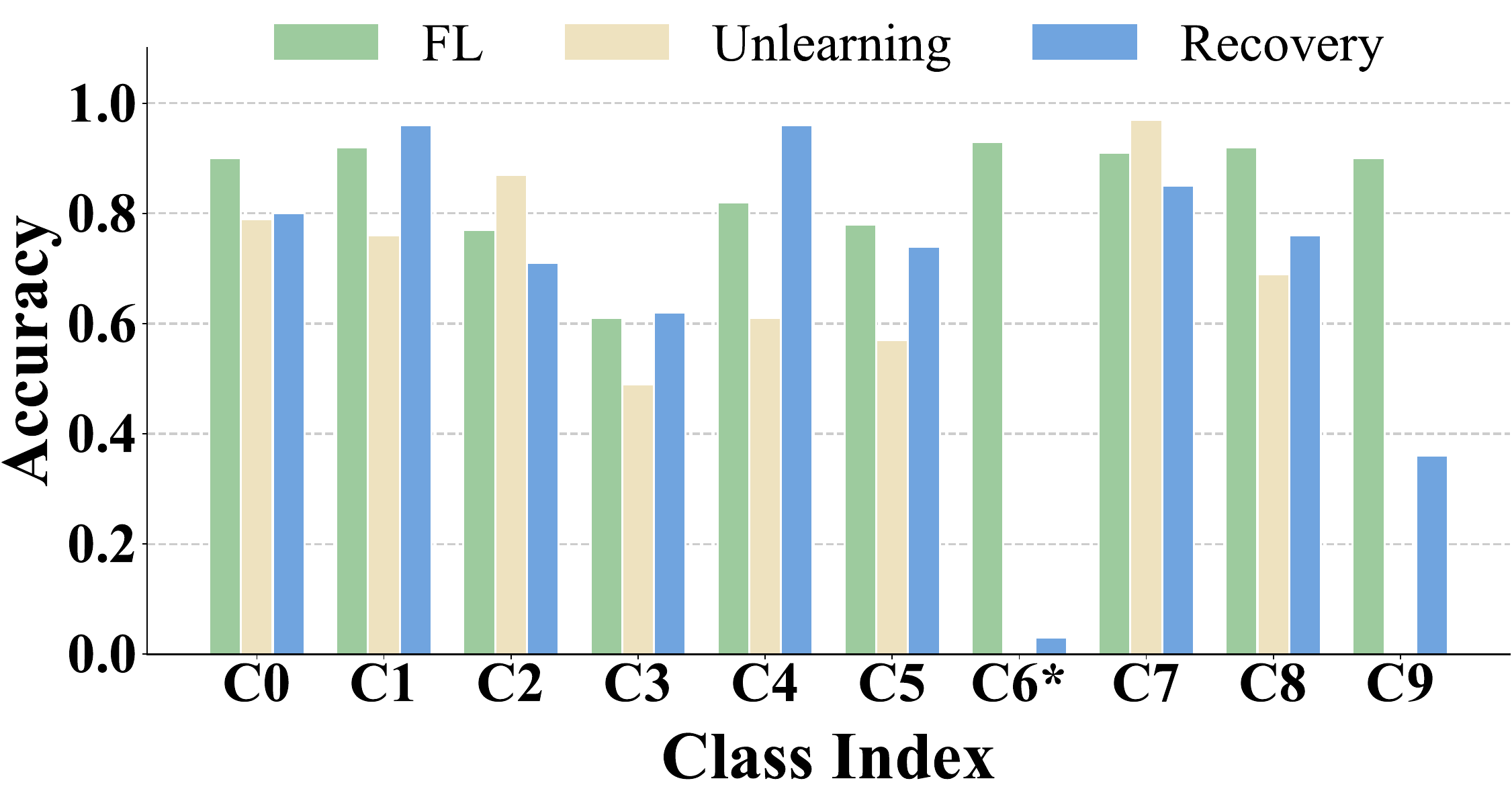}
        \caption{FedOSD}
        \label{fig:acc_class_FedOSD}
    \end{subfigure}
    \hfill 
    \begin{subfigure}[b]{0.3\textwidth}
        \centering
        \includegraphics[width=\linewidth]{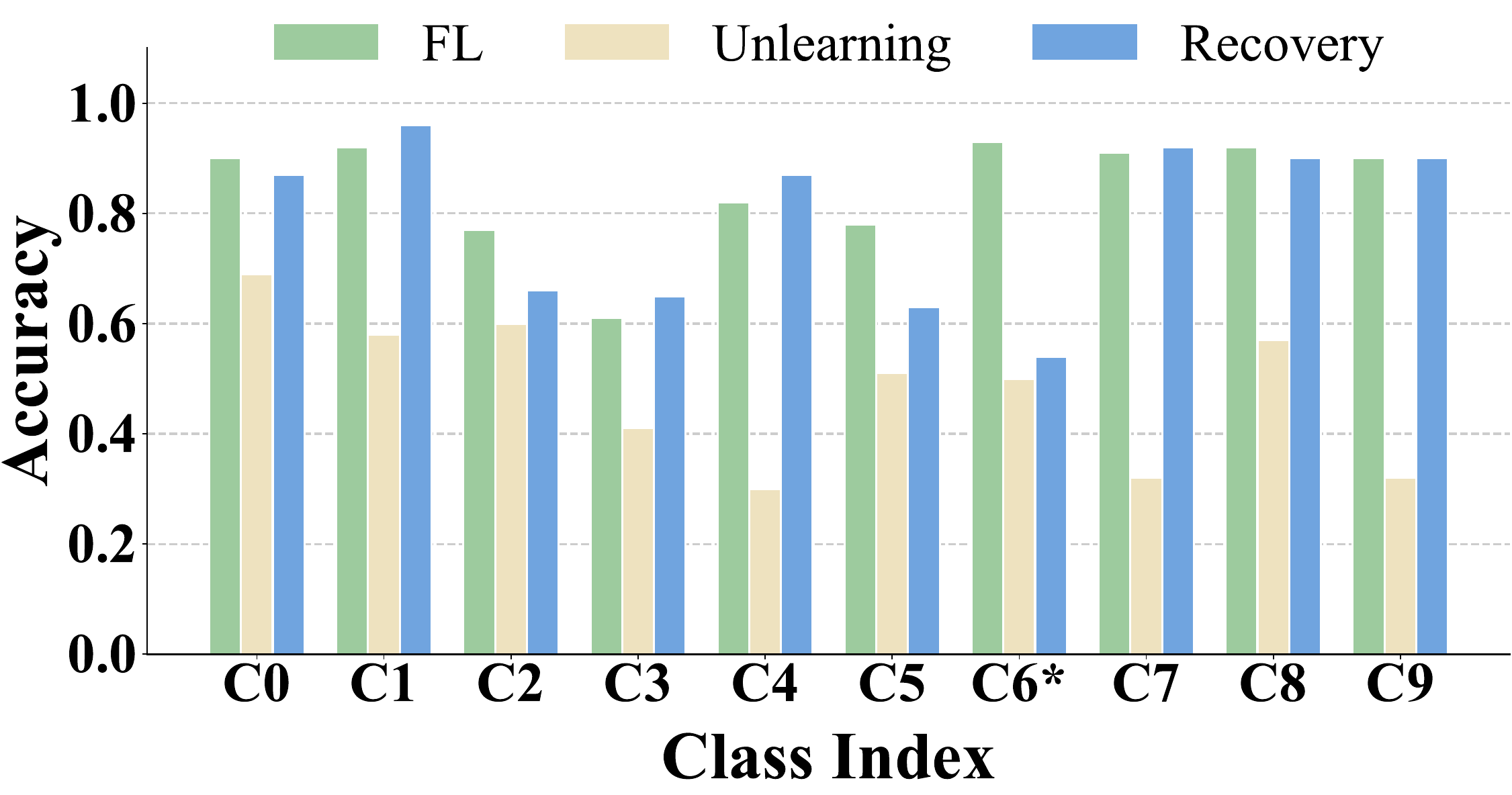}
        \caption{NoT}
        \label{fig:acc_class_NoT}
    \end{subfigure}
    \hfill 
    \begin{subfigure}[b]{0.3\textwidth}
        \centering
        \includegraphics[width=\linewidth]{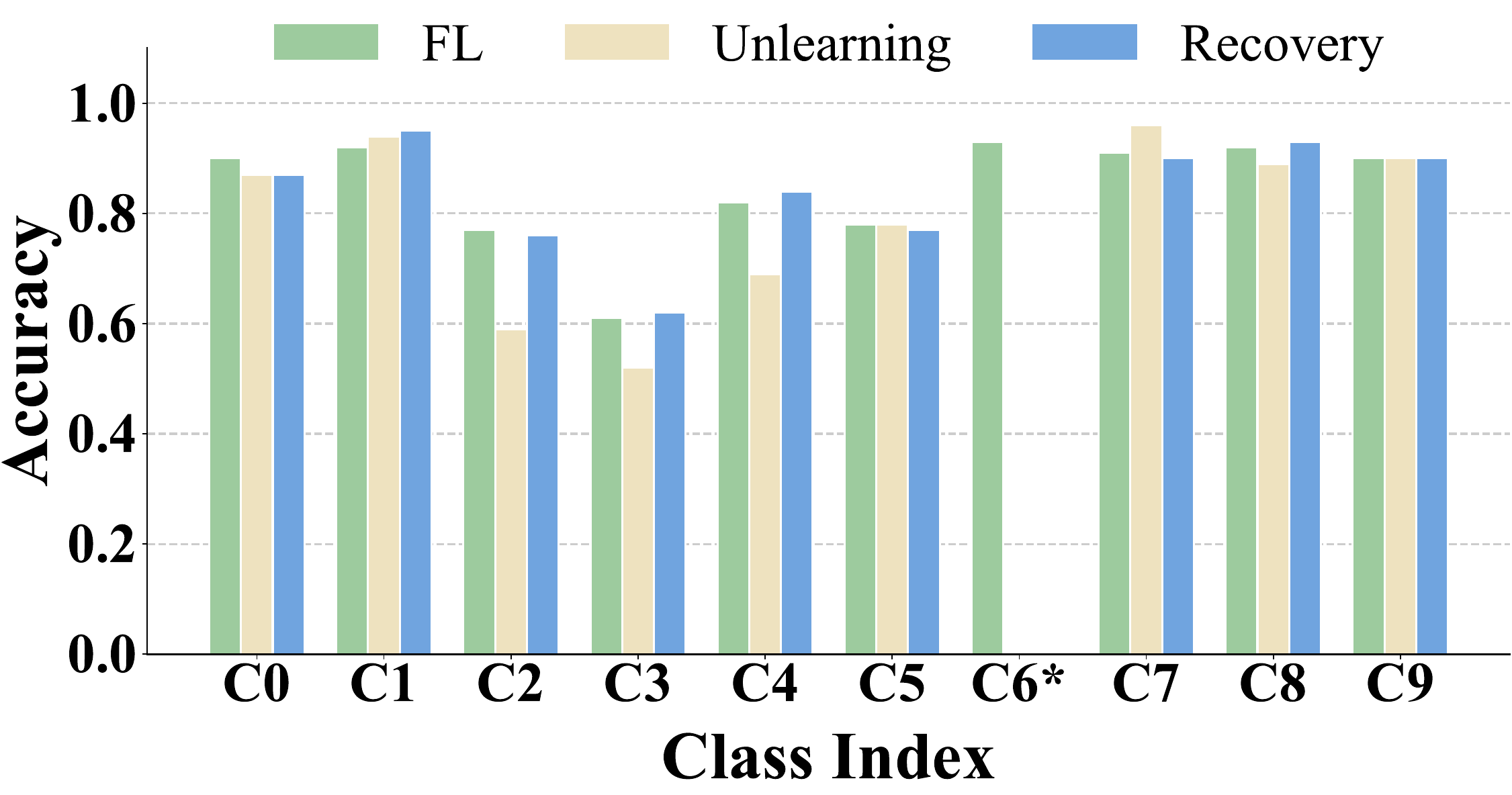}
        \caption{FedCARE}
        \label{fig:acc_class_FedCARE}
    \end{subfigure}
    
    \caption{{Class-level unlearning performance across different classes on CIFAR-10 (non-IID).} The ($*$) denotes the class to be forgotten.}
    \label{fig:class_wise_accuracy}
\end{figure*}

\begin{table*}[t!]
    \centering
    \scriptsize
    \setlength{\tabcolsep}{8pt}
    \renewcommand{\arraystretch}{0.7}
        \begin{tabular}{llcccccccc}
            \toprule
            \multirow{2}{*}{} & 
            \multirow{2}{*}{\textbf{Method}} & 
            \multicolumn{4}{c}{\textbf{non-IID}} & 
            \multicolumn{4}{c}{\textbf{IID}} \\
            \cmidrule(lr){3-6} \cmidrule(lr){7-10}
             & & \textbf{U-Acc} & \textbf{R-ACC} & \textbf{Time} ($\downarrow$) & \textbf{FLOPs} ($\downarrow$) & \textbf{U-Acc} & \textbf{R-ACC} & \textbf{Time} ($\downarrow$) & \textbf{FLOPs} ($\downarrow$) \\
            \midrule
            
            \multirow{5}{*}{\shortstack[l]{CIFAR-10}} 
            & Retrain & 0 & 83.96\% & 8587.48s & $7.23 \times 10^{15}$ & 0 & 84.69\% & 7291.03s & $6.02 \times 10^{15}$ \\
            \cmidrule(lr){2-10} 
            
            & MoDe & 0 & 76.59\% & 3105.61s & $1.36 \times 10^{15}$ & 1.57\% & 77.20\% & 3000.54s & $1.12 \times 10^{15}$ \\

            & FedOSD & 2.76\% & 81.77\% & 483.28s & $8.17 \times 10^{14}$ & 3.54\% & 82.77\% & 477.31s & $8.05 \times 10^{14}$ \\
            
            & NoT & 54.10\% & 78.53\% & 6969.16s & $6.02 \times 10^{15}$ & 33.90\% & 85.83\% & 6143.88s & $5.42 \times 10^{15}$ \\
            
            & {FedCARE} & {0} & {83.93\%} & {300.53s} & ${7.45 \times 10^{14}}$ & {0} & {83.34\%} & {280.87s} & ${7.53 \times 10^{14}}$ \\
            \bottomrule
        \end{tabular}
    
    \caption{{Class-level unlearning performance.} Class 6 is removed from all clients.}
    \label{tab:class_UL}
\end{table*}

\begin{table}[t!]
    \centering
    \scriptsize
    \setlength{\tabcolsep}{3pt}
    \renewcommand{\arraystretch}{0.7}
    \begin{tabular}{llcccccc}
        \toprule
        \multirow{2}{*}{} & 
        \multirow{2}{*}{\textbf{Method}} & 
        \multicolumn{3}{c}{\textbf{Accuracy}} & 
        \multicolumn{1}{c}{\textbf{Privacy}} & 
        \multicolumn{2}{c}{\textbf{Cost}} \\
        \cmidrule(lr){3-5} \cmidrule(lr){6-6} \cmidrule(lr){7-8}
         & & \textbf{R-Acc} & \textbf{U-Acc}  & \textbf{Test}  & \textbf{MIA} ($\downarrow$) & \textbf{Time} ($\downarrow$) & \textbf{FLOPs} ($\downarrow$) \\
        \midrule
        
        \multirow{4}{*}{\rotatebox{90}{\shortstack{CIFAR-10}}} 
        & Retrain & 82.52\% & 80.79\% & 82.08\% & 50.71\% & 8696.20s & $6.21 \times 10^{15}$ \\
        \cmidrule(lr){2-8} 
        
        & FedOSD & 84.29\% & 79.04\% & 82.13\% & 54.38\% & 517.14s & $9.61 \times 10^{14}$ \\
        & NoT & 80.07\% & 78.36\% & 79.10\% & 58.65\% & 3780.26s & $3.31 \times 10^{15}$ \\

        & {FedCARE} & {82.21\%} & {80.50\%} & {81.56\%} & {51.38\%} & {137.79s} & ${1.65 \times 10^{14}}$ \\
        \bottomrule
    \end{tabular}
    \caption{{Instance-level unlearning performance under non-IID settings.} One client is randomly selected to remove 10\% of its local samples. IID scenarios are omitted since removing a small fraction of data under homogeneous distributions has negligible impact on the global model.}
    \label{tab:instance_FL}
\end{table}

\paragraph{Unlearning Effectiveness and Utility Maintenance}
We evaluate utility maintenance across three granularities: client, instance, and class levels. For client-level unlearning, Table~\ref{tab:client_UL} shows that FedCARE achieves a U-Acc of 78.52\% on CIFAR-10, closely matching the Retraining baseline (78.63\%), indicating effective disentanglement of client-specific and shared knowledge. In contrast, partition-based and distillation-based methods, such as FedEraser and MoDe, suffer concurrent drops in both U-Acc and R-Acc, reflecting limited generalization. For instance-level unlearning (Table~\ref{tab:instance_FL}), FedCARE maintains an R-Acc of 82.21\% on CIFAR-10, comparable to Retraining (82.52\%), demonstrating minimal collateral impact on global decision boundaries. For class-level unlearning, Table~\ref{tab:class_UL} shows that FedCARE uniquely achieves complete erasure, driving U-Acc to 0\% under both IID and non-IID settings while preserving R-Acc close to Retraining. By contrast, FedOSD retains residual U-Acc under both IID and non-IID settings (3.54\% and 2.76\%), and NoT preserves substantial accuracy (54.10\%), indicating ineffective removal of the target concept. Figures~\ref{fig:client_wise_accuracy} and~\ref{fig:class_wise_accuracy} further confirm that FedCARE incurs minimal utility degradation on retained classes during both unlearning and recovery. Since FedOSD approximates class-level unlearning via client-level unlearning, it exhibits similar performance trends across Figures~\ref{fig:acc_client_FedOSD} and~\ref{fig:acc_class_FedOSD} under identical data partitions.

\begin{figure}[t]
    \centering
    \renewcommand{\arraystretch}{0.85}
    \includegraphics[width=0.65\linewidth]{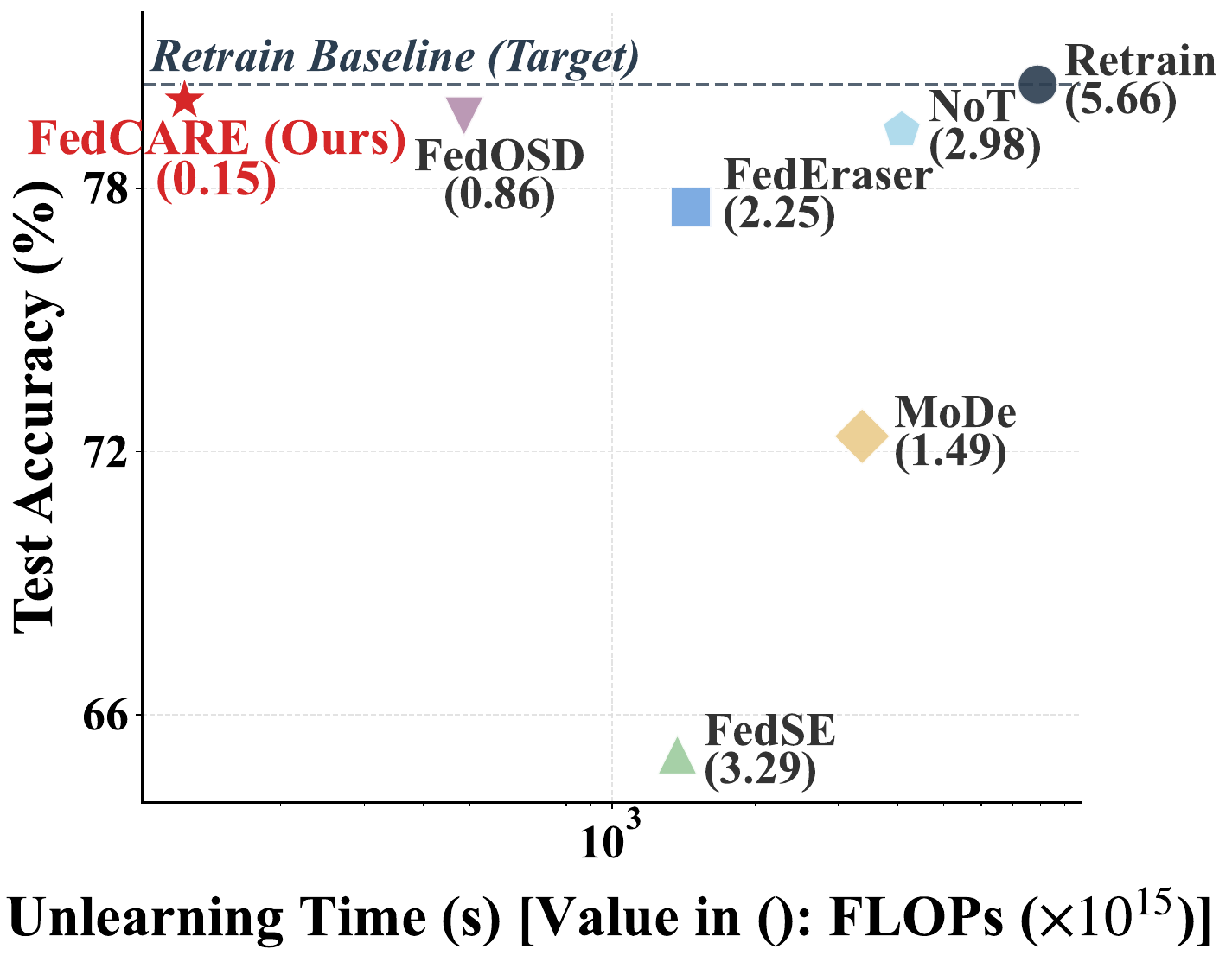}
    \caption{{Efficiency comparison between FedCARE and baseline methods.}}
    \label{fig:efficiency_time}
\end{figure}

\paragraph{Efficiency and Scalability Analysis}
FedCARE demonstrates strong efficiency in both execution time and computational overhead (Figure~\ref{fig:efficiency_time}). As shown in Table~\ref{tab:client_UL}, FedCARE completes CIFAR-10 unlearning in 125.39s, yielding a $63\times$ speedup over Retrain and a $32\times$ speedup over NoT (4078s), which incurs high latency due to iterative fine-tuning for repairing disruptive weight negation. Beyond wall-clock time, the frozen-backbone recovery strategy significantly reduces computation, limiting FLOPs to $1.49 \times 10^{14}$. This is substantially lower than the compute-intensive FedOSD ($8.65 \times 10^{14}$) and an order of magnitude more efficient than storage-heavy methods such as FedSE, highlighting FedCARE’s suitability for resource-constrained edge environments.


\paragraph{Privacy and Security Verification}
We evaluate data privacy using MIA and model security using backdoor defense. As shown in Table~\ref{tab:instance_FL} (and Table~\ref{tab:client_UL} for client-level unlearning), FedCARE achieves an instance-level MIA score of 50.38\% on CIFAR-10, which is close to random guessing, indicating that membership information of forgotten samples is effectively removed. In contrast, NoT and FedEraser exhibit higher MIA scores ($>57\%$), suggesting that identifiable traces remain in the model. For backdoor defense (Table~\ref{tab:backdoor_results}), FedCARE reduces the ASR to 0\% while maintaining high utility (79.42\%). Although FedOSD also eliminates the backdoor, it incurs substantially higher computational cost, whereas MoDe and NoT fail to fully remove embedded triggers, with ASR exceeding 25\%, which poses serious security risks and limits their applicability in safety-critical scenarios.

\begin{table}[t!]
    \centering
    \scriptsize
    \setlength{\tabcolsep}{6pt}
    \renewcommand{\arraystretch}{0.7}
    
    \begin{tabular}{llcccc}
        \toprule
        \multirow{2}{*}{} & 
        \multirow{2}{*}{\textbf{Method}} & 
        \multicolumn{1}{c}{\textbf{Accuracy}} & 
        \multicolumn{1}{c}{\textbf{Security}} & 
        \multicolumn{2}{c}{\textbf{Cost}} \\ 
        \cmidrule(lr){3-3} \cmidrule(lr){4-4} \cmidrule(lr){5-6}
         &  & \textbf{Test} & \textbf{ASR} ($\downarrow$)& \textbf{Time} ($\downarrow$) &\textbf{FLOPs} ($\downarrow$) \\
        \midrule
        
        \multirow{7}{*}{\rotatebox{90}{\shortstack{CIFAR-10}}}
   
        & Retrain & 80.01\% & 0& 7892.47s &$5.66 \times 10^{15}$ \\
        \cmidrule(lr){2-6} 
        & FedEraser  & 78.33\% & 12.56\%&1429.02s& $2.17 \times 10^{15}$ \\
        & FedSE  & 62.72\%& 8.71\% &1380.71s& $3.43 \times 10^{15}$ \\
        & MoDe  & 73.45\%& 25.87\% & 3437.68s&$1.72 \times 10^{15}$ \\
        & FedOSD& 79.31\% & 0  &479.26s& $8.65 \times 10^{14}$ \\
        & NoT  & 77.66\%& 38.64\% & 4022.94s&$2.77 \times 10^{15}$ \\

        & {FedCARE}  & {79.42\%}& {0} & 200.37s&${1.89 \times 10^{14}}$ \\
        \bottomrule
    \end{tabular}
    \caption{{Backdoor defense performance under non-IID settings.}}
    \label{tab:backdoor_results}
\end{table}

    

\begin{table}[t!]
    \centering
    \scriptsize
    \setlength{\tabcolsep}{6pt}
    \renewcommand{\arraystretch}{0.7}
    \begin{tabular}{lcccccc}
        \toprule
        \multirow{2}{*}{\textbf{Method}} 
        & \multicolumn{3}{c}{\textbf{Accuracy}} 
        & \multicolumn{1}{c}{\textbf{Privacy}} 
        & \multicolumn{1}{c}{\textbf{Security}} \\
        \cmidrule(lr){2-4} \cmidrule(lr){5-5} \cmidrule(lr){6-6}
        & \textbf{R-ACC} 
        & \textbf{U-ACC} 
        & \textbf{Test} 
        & \textbf{MIA ($\downarrow$)} 
        & \textbf{ASR ($\downarrow$)} \\
        \midrule
        Retrain  & 81.62\% & 78.60\% & 80.01\% & 50.20\% & 0.00\%\\
        \midrule
        {M1}  & 80.37\% & 79.44\% & 79.06\% & 51.36\% & 33.50\%\\
        {M2}  & 77.20\% & 67.00\% & 75.44\% & 50.38\% & 0.00\%\\
        {M3}  & 81.66\% & 77.54\% & 76.79\% & 57.12\% & 45.77\%\\
        {M4}  & 82.74\% & 80.70\% & 79.95\% & 55.35\% & 29.60\%\\
        {M5}  & 81.16\% & 79.82\% & 78.04\% & 51.67\% & 20.71\% \\
        {FedCARE}  
        & {81.30\%} & {50.38\%} & {78.42\%} & {78.52\%} & {0.00\%}\\
        \bottomrule
    \end{tabular}
    \caption{{Ablation study on client-level unlearning.} Performance comparison of different components under the non-IID setting.}
    \label{tab:ablation_cifar10}
\end{table}

\subsection{Ablation Experiments}

To rigorously assess the contribution of each component in FedCARE, we conduct ablation studies on the non-IID CIFAR-10 dataset using ResNet-18. Specifically, we consider five variants: (M1) replacing Group Normalization in the generator with Batch Normalization to examine the effect of normalization under heterogeneous data distributions; (M2) performing unlearning via raw gradient ascent on private data without the projection constraint or protective gradient to evaluate the necessity of conflict-aware projection; (M3) applying standard FedAvg for post-unlearning recovery without any anti-relearning mechanism to expose rollback risks; (M4) disabling backbone freezing during recovery while retaining server-side projection to test whether global calibration alone can prevent feature reconstruction; and (M5) removing server-side vector filtering while keeping local backbone freezing to examine whether local constraints alone suffice to suppress directional rollback of the classifier head.



As shown in Table~\ref{tab:ablation_cifar10}, the ablation results validate the necessity of each component in FedCARE. During unlearning, removing the projection constraint (M2) yields the lowest R-ACC (77.20\%), confirming that unconstrained gradient ascent severely degrades model utility, while replacing Group Normalization with Batch Normalization in the generator (M1) harms both utility and security, leading to a high ASR of 33.50\% and highlighting the importance of constructing a reliable generic knowledge subspace under non-IID settings. During recovery, the FedAvg baseline (M3) exhibits severe relearning with an ASR of 45.77\%, and disabling local backbone freezing (M4) causes a substantial ASR rebound to 36.60\%, indicating that the backbone is a primary carrier of backdoor features and that server-side projection alone is insufficient to prevent reconstruction. Although retaining backbone freezing but removing global vector filtering (M5) reduces ASR to 20.71\%, residual directional drift remains. In contrast, FedCARE combines backbone freezing with vector filtering to achieve complete backdoor removal (0\% ASR) while maintaining strong utility (81.30\% R-ACC), demonstrating the effectiveness of the proposed dual-constraint recovery mechanism.


\section{Conclusion}

This paper presents FedCARE, a unified and low-overhead federated unlearning framework that addresses key challenges in FU, including high unlearning cost, utility degradation, and unintended relearning during recovery. By combining conflict-aware projected unlearning with a relearning-resistant recovery mechanism, FedCARE enables effective removal of client-, instance-, and class-level information without retraining from scratch. Extensive experiments across multiple datasets and model architectures under both IID and non-IID settings demonstrate that FedCARE achieves reliable forgetting, strong utility retention, and improved robustness compared to state-of-the-art FU methods. These results indicate that FedCARE provides a practical solution for privacy-compliant federated learning systems.



\FloatBarrier
\bibliographystyle{named}
\bibliography{ijcai26}

\clearpage
\appendix

\end{document}